\documentclass[review,OA]{AAAI-Std}
\usepackage{graphicx}
\def\artid{0001}%
\usepackage[T1]{fontenc}
\usepackage{bm}
\usepackage{amsmath}
\usepackage{xcolor}
\usepackage{subfigure}
\usepackage{multirow}
\usepackage{caption}
\usepackage{makecell}
\usepackage{array}
\newcolumntype{L}[1]{>{\raggedright\let\newline\\\arraybackslash\hspace{0pt}}m{#1}}
\newcolumntype{C}[1]{>{\centering\let\newline  \\\arraybackslash\hspace{0pt}}m{#1}}
\newcolumntype{R}[1]{>{\raggedleft\let\newline \\\arraybackslash\hspace{0pt}}m{#1}}

\begin{document}

\title{Beyond Scaleup: Knowledge-aware Parsimony Learning
	from Deep Networks}


\author{Q\textsc{uanming} Y{ao}}
\author{Y\textsc{ongqi} Z\textsc{hang}}
\author{Y\textsc{aqing} W\textsc{ang}}
\author{N\textsc{an} Y\textsc{in}}
\author{J\textsc{ames} K\textsc{wok}}
\author{Q\textsc{iang} Y\textsc{ang}}

\address{Q.~Yao is an Assistant Professor in the Department of Electronic Engineering at the Tsinghua University.
Y.~Zhang is an Assistant Professor in the Department of Data Science and Analytics Thrust at the Hong Kong University of Science and Technology (Guangzhou).
Y.~Wang is an Associate Professor in Beijing Institute of Mathematical Sciences and Applications. 
N.~Yin is a Research Associate in the Department of Computer Science and Engineering at the Hong Kong University of Science and Technology.
J.~Kwok and Q.~Yang are Professors in the Department of Computer Science and Engineering at the Hong Kong University of Science and Technology.}

\address{Correspondence is to Q. Yao
	at \url{qyaoaa@tsinghua.edu.cn}}

\abstract[Abstract]{
The brute-force scaleup of training datasets,
learnable parameters 
and computation power, 
has become a prevalent strategy for developing more
robust learning models. However, due to bottlenecks in data, computation, and
trust, the sustainability of this strategy is a serious concern. 
In this paper, 
we attempt to address this issue in a parsimonious manner (i.e., achieving greater potential with simpler
models). The key is to drive models using domain-specific knowledge, such as
symbols, logic, and formulas, instead of purely relying on scaleup. 
This approach
allows us to build a framework that uses this knowledge as “building blocks” to
achieve parsimony in model design, training, and interpretation. Empirical results
show that our methods surpass those that typically follow the scaling law. We
also demonstrate our framework in AI for science, specifically
in the problem of drug-drug interaction prediction. We hope our research can
foster more diverse technical roadmaps in the era of foundation models.
}


\maketitle

\section{Introduction}

The learning techniques have progressed from manual feature engineering to shallow models, 
then to deep networks, and now to foundation models, 
achieving great success in the field of computer vision, 
natural language understanding
and speech processing. 
Specifically, large language models, like ChatGPT~\citep{ouyang2022training}, 
as representatives of foundation model, has shown strong performance in versatile learning, 
which can adopted in many different tasks. 
The belief is that larger models can be more expressive, 
thus are likely to generalize better given sufficient training data~\citep{Jordan2015,Donoho2017}. 
This gives birth to the current roadmap, i.e., 
achieving stronger performance by aggressively scaleup the size of data and models, 
which is also observed as scaling law~\citep{kaplan2020scaling}.

\begin{figure*}[t]
    \centering
    \includegraphics[width=\textwidth]{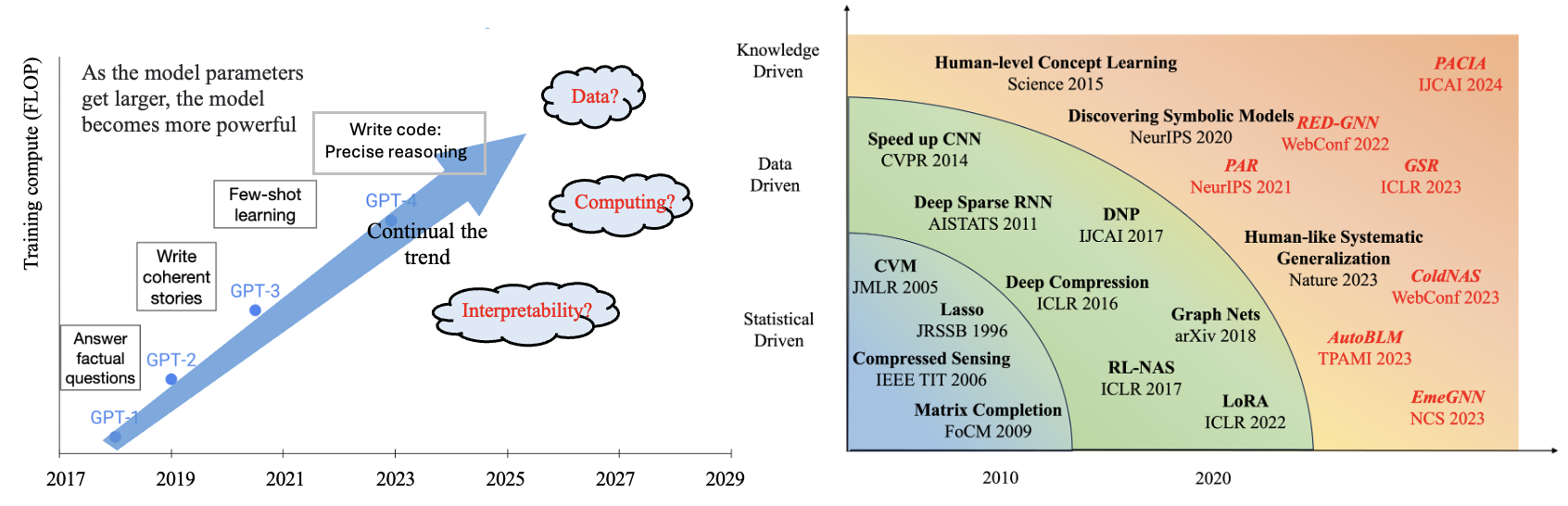}
    
    \vspace{-10px}
    \caption{The data, computational and trust bottlenecks of LLMs (left), and the development of parsimony learning (right).}
    \label{fig:challenge}
    \vspace{-15px}
\end{figure*}

However, 
such a roadmap potentially leads to serious problems 
(as shown in the left of Figure~\ref{fig:challenge}): 
\textit{Data bottleneck}—the scaling law relies on vast amounts of high-quality data, 
yet all available online corpora are projected to be exhausted by 2028~\citep{villalobosposition};
\textit{Computational bottleneck}—the exponentially growing number of parameters demands substantial high-performance computing power, yet the pace of hardware development struggles to keep up with this rapid increase~\citep{wolters2024memory};
and 
\textit{Trust bottleneck}—the scaling law follows the data driven path, disregarding internal logical relationships, which leads to opaque reasoning processes and severe hallucination issues~\citep{duan2024llms}. 
These indicate that the current roadmap is not sustainable, 
and motivates us to ask:
where is the way if scaleup fails?

\begin{figure*}[t] 
	\centering
	\includegraphics[width=0.95\textwidth]{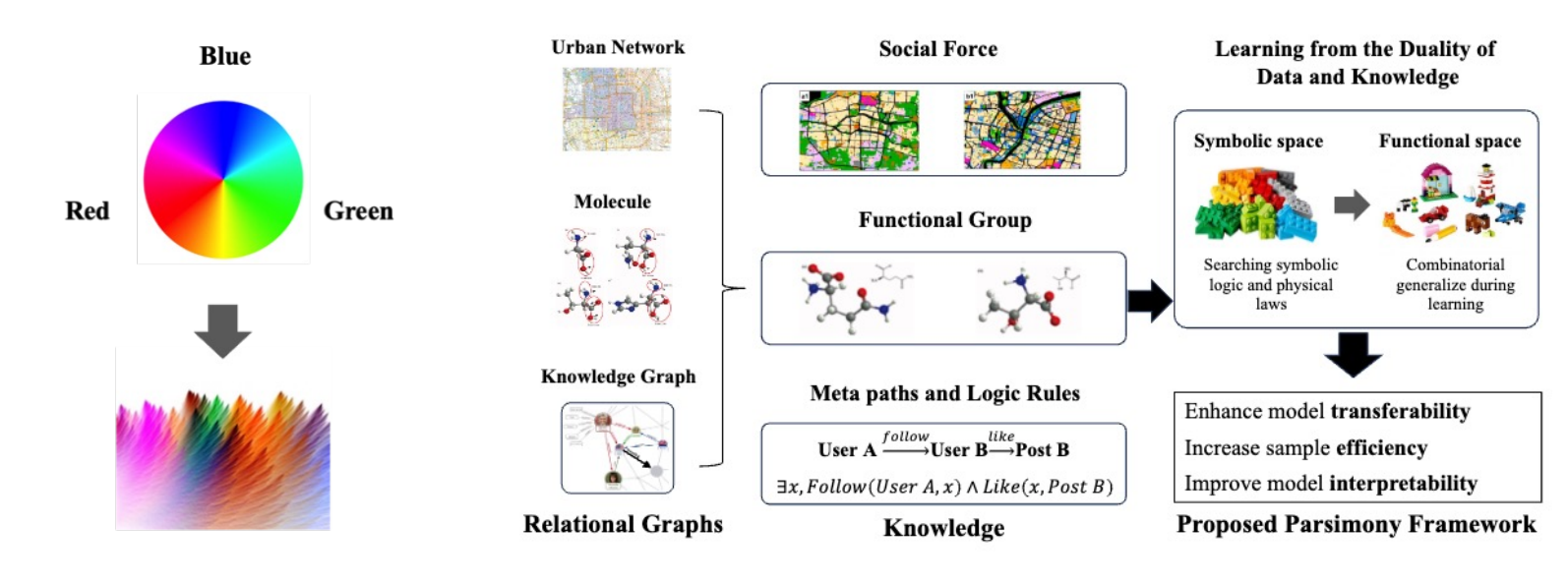}
	
	\vspace{-15px}
	\caption{The three primary colors (left) and the knowledge-aware parsimony learning framework (right).}
	\label{problem}
	\vspace{-10pt}
\end{figure*}

To address this question, we look back to the fundamental principles that drive machine learning. 
Albert Einstein famously stated, ``Everything should be made as simple as possible, but not simpler.'' 
This insight has inspired many AI researchers to develop learning techniques that embrace parsimony, 
aiming to achieve ``maximum output with minimal input'' and ``leveraging small inputs for significant effects''.
Traditional approaches to achieving parsimony learning have typically followed statistical-driven and data-driven methods
(as shown in the right of Figure~\ref{fig:challenge}). 
Statistical driven methods are rooted in well-established statistical theories and principles.
Example works are
core vector machines~\citep{tsang2005core},
compressed sensing~\citep{donoho2006compressed},  
matrix completion~\citep{candes2012exact}
and Lasso~\citep{tibshirani1996regression}.
These methods penalize complexity and encourage models to focus on essential features. 
Later on,
data driven methods gradually developed. 
Example works are
Deep Sparse CNN\citep{glorot2011deep}, Speedup CNN~\citep{jaderberg2014speeding}, Deep Compression~\citep{han2015deep}, RL-NAS~\citep{zoph2016neural}, DNP~\citep{Liu2017}
and LoRA~\citep{hu2021lora}.
These methods model and make predictions based on the data, 
rather than relying on explicit statistical assumptions. 
However, the development of data driven approaches would
still lead to the way of scaling law, which suffers from the bottlenecks mentioned above.

Intuitively, humans accumulate knowledge in the form of symbols, concepts, rules, and principles, 
which allows us to learn a wide range of subjects or skills, apply them across various tasks, 
quickly adapt to new tasks with few or no demonstrations, 
and uncover the underlying reasons behind different phenomena.
Inspired by \citep{lake2015human,cranmer2020discovering,lake2023human}, 
which shows that neural networks can achieve human-like systematicity with smaller models and achieve performance comparable to that of larger models, 
we propose knowledge driven methods that emphasize guiding machine learning models with domain-specific knowledge, 
using it as “building blocks” to achieve parsimony in learning. This approach allows us to tackle the bottlenecks of scaleup, 
focusing instead on efficient model design, training, and interpretation. 
We demonstrate the effectiveness of this framework in AI for science, 
particularly in addressing the problem of drug-drug interaction prediction. 
Empirical results show that our methods can outperform those limited by the scaling law.

\section{Research Landscape}

Our research landscape starts from relational graphs.
Different from images, natural language and speech, 
relational graphs are a way to represent knowledge using nodes and edges, 
which symbolizes human knowledge in a structured form.
The application of graphs are widely spread in 
real-life scenarios, such as 
designing and
planning of
urban networks, 
prediction of molecular properties, 
reasoning from knowledge graphs, 
and recommendation systems
\cite{wang2017knowledge,wu2020comprehensive}.

Our key innovation lies in enabling parsimony learning through knowledge-aware approaches. 
Specifically, knowledge, i.e., symbolic logic and physical laws, is ubiquitous in real-world scenarios and coexists with data. 
As illustrated on the left side of Figure~\ref{problem}, 
the diverse colors found in nature can be derived from three primary colors (i.e., red, blue, and green). 
Building on this, we propose the concept of the ``duality of knowledge and data'', 
which posits that data is both numerical and knowledge-based.
We subsequently design a framework capable of learning from both data and knowledge simultaneously, 
thereby effectively achieving parsimony in learning. 
As shown on the right side of Figure~\ref{problem}, the framework first identifies symbolic logic and physical laws at the knowledge level, 
then performs combinatorial generalization of this knowledge at the data level. By treating knowledge as ``building blocks'' and focusing on its learning and generalization, the framework leverages simple knowledge to solve complex problems.

\begin{figure*}[t] 
	\centering
	\includegraphics[width=1\textwidth]{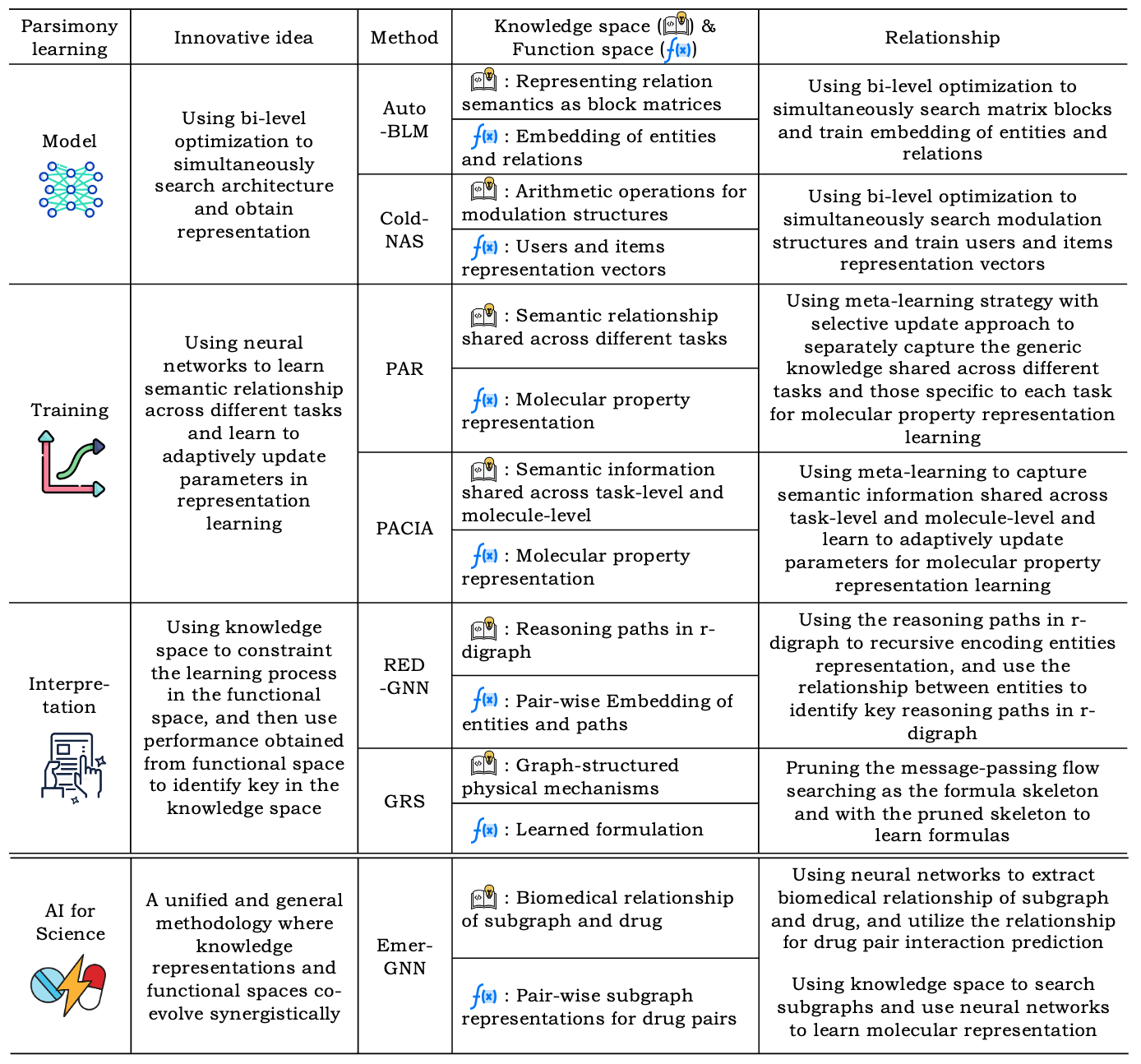}
	
	\vspace{-10px}
	\captionof{table}{Illustration of knowledge space, function space, and the relationship in each method.}
	\label{tab:total}
	\vspace{-10pt}
\end{figure*}

Here, 
we incorporate domain-specific knowledge into our framework and propose parsimony on model to overcome computational bottlenecks, 
parsimony on training to mitigate data bottlenecks, 
and parsimony on interpretation to address trust bottlenecks. We then explore the potential of our approach in AI for Science, 
aiming to tackle all these bottlenecks within a unified framework.
As shown in 
Table~\ref{tab:total}, we illustrate the knowledge space, function space, and their relationships in each part.

\begin{itemize}
\item
\textit{Parsimony on Model}. 
The goal of architectural parsimony is 
to utilize simple architecture to achieve the comparable performance of complex models.
As introduced in \citep{lake2023human}, 
simple neural networks can achieve human-like systematicity when optimized for compositional skills, 
offering the potential for smaller models to achieve performance comparable to that of larger ones. We use bi-level optimization to simultaneously search architecture in knowledge space and obtain representation in functional space to achieve parsimony on model.
Following this intuition, we propose AutoBLM \citep{zhang2022bilinear} and ColdNAS \citep{Wu2023}, 
which utilize prior knowledge to streamline architectures.
By achieving this, we can maintain high performance with simpler structures, thereby efficiently addressing computational bottlenecks.

\vspace{5px}
\item 
\textit{Parsimony on Training}. 
The purpose of training parsimony is to reduce the 
number of parameters needs to be updated during the training.
Inspired by~\citep{thrun1998learning,rastogi2022learning,ilharco2022editing}, 
which shows that task-related knowledge 
help to determine parameters by simple arithmetic operations
without training,
we propose new approaches for learning parsimony~\citep{wang2021property,yao2024property,wu2024pacia}. Specifically, we use neural networks to learn semantic relationships across different tasks in knowledge space and learn to update parameters in representation learning in function space adaptively.
These approaches leverage 
knowledge to guide the fine-tuning process on specific tasks under few-shot circumstances. 
By achieving this, we can efficiently utilize knowledge to guide the optimization of related tasks with limited training data, thereby efficiently addressing data bottlenecks.

\vspace{5px}
\item 
\textit{Parsimony on Interpretation}. 
The parsimony on interpretation is to identify important evidence in face to massive connections on graphs.
Our primary approach is to encapsulate logical rules within graphs as supporting evidence, 
utilizing the logical interpretation of knowledge to 
elucidate the reasoning processes 
of models~\citep{shi2023learning,zhang2022knowledge}. We use knowledge space to constraint the learning process in the functional space and then use performance obtained from functional space to identify keys in the knowledge space.
By achieving this, we can efficiently interpret the model's result with 
subgraph, 
thereby efficiently addressing trust bottlenecks.

\vspace{5px}
\item 
\textit{Potential in AI for Science}. 
To efficiently co-evolve the knowledge and functional spaces to address the bottlenecks of data, computation, and interoperability, we propose a unified framework specifically designed for drug development.
To enhance the interpretability of drug-drug interactions, we employ neural networks to extract biomedical relationships between subgraphs and drugs within the knowledge space. These relationships are then utilized for drug-pair interaction prediction within the functional space. Additionally, by leveraging simple architectures (subgraphs) identified in the knowledge space and neural networks for learning molecular representations, we achieve superior performance compared to traditional methods. 
Specifically,
we take drug-drug interaction as an application, which predicts interactions between emerging and existing drugs. By extracting path-based subgraphs and learning subgraph representation, we can efficiently reduce the data and computational requirements and identify evidence with subgraphs. 
\end{itemize}

\begin{figure}[t]
	\centering
	\includegraphics[width=0.9\linewidth]{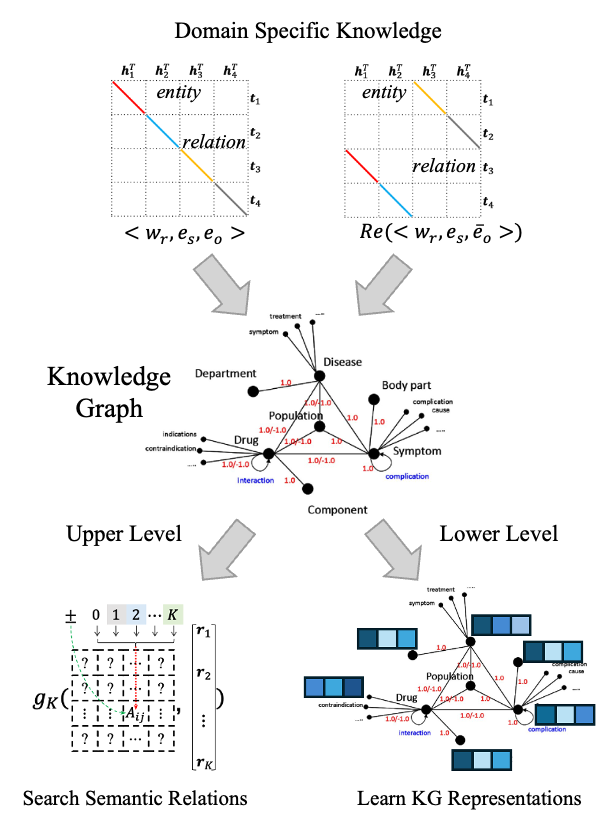}
	
	\vspace{-10pt}
	\caption{AutoBLM first sets up a search space by analyzing existing scoring functions and then utilizes the bi-level optimization to extract semantics and relationships simultaneously.}
	\label{fig:autoblm}
	\vspace{-20pt}
\end{figure}

\section{Parsimony on Model}

The goal of parsimony on model is to match the performance of complex models using a simpler architecture. 
To achieve this, it is necessary to prune the network architecture or search the simple structure to guide the model simplification or semantic reorganization. Inspired by~\citep{lake2023human}, 
which highlights the potential of achieving superior performance with standard neural network architectures through ``knowledge-aware'' meta-learning, 
we propose approaches to architectural parsimony by leveraging knowledge as prior information 
to guide the search for alternative but simpler structures in learning.

\begin{figure}[t]
	\centering
	\includegraphics[width=0.48\textwidth]{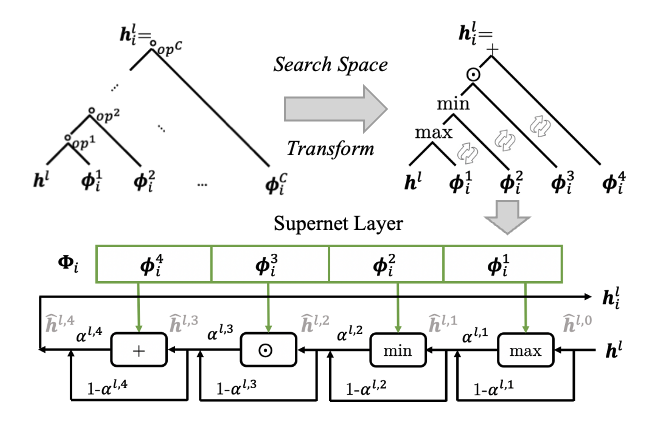}
	
	\vspace{-10pt}
	\caption{ColdNAS uses a hypernetwork to map each user’s history interactions to user-specific parameters which are then used to modulate the predictor, and formulate how to modulate and where to modulate as a NAS problem.}
	\label{fig:coldnas} 
	\vspace{-10pt}
\end{figure}

\subsection{Automated Bi-linear Scoring Function Design}

In knowledge graph learning, the scoring function is a key component that measures the plausibility of edges \citep{wang2017knowledge}. 
Human experts often design various complex scoring functions to evaluate edge plausibility, 
which can be redundant and inefficient in model design. 
To achieve parsimony across different scoring functions, we propose AutoBLM~\cite{zhang2022bilinear}, 
which focuses on semantics and leverages computational power to reconfigure simple architectural elements, 
enabling efficient learning across diverse tasks. 
Specifically, AutoBLM employs a bi-level framework to simultaneously extract 
semantic representations (i.e., the embeddings of entities and relations in knowledge graph) and search the relations of entities, 
which is shown in Figure~\ref{fig:autoblm}. 
The lower level learns representations from the training dataset, 
while the upper level searches the knowledge space to discover semantic relationships in a unified search space, using the validation dataset. 
Additionally, AutoBLM enables the learning of new models that are better adapted to specific datasets through an efficient evolutionary search algorithm, enhancing transferability.

The performance of AutoBLM 
is evaluated with 
knowledge graph reasoning benchmarks.
The results,
evaluated with mean reciprocal ranking (MRR) metric,
are shown in 
Table~\ref{tab:autoblm}.
Compared with the complex neural network models
Interstellar \citep{zhang2020interstellar} and CompGCN \citep{vashishth2020composition},
AutoBLM gains significant improvement on the three datasets.
In particular,
it recombines the simple relation to simulate complex structure
and outperforms the
deep neural network models, which often have better expressiveness with higher computational costs.

\begin{figure*}[t]
	\centering	
	\includegraphics[width=1\textwidth]{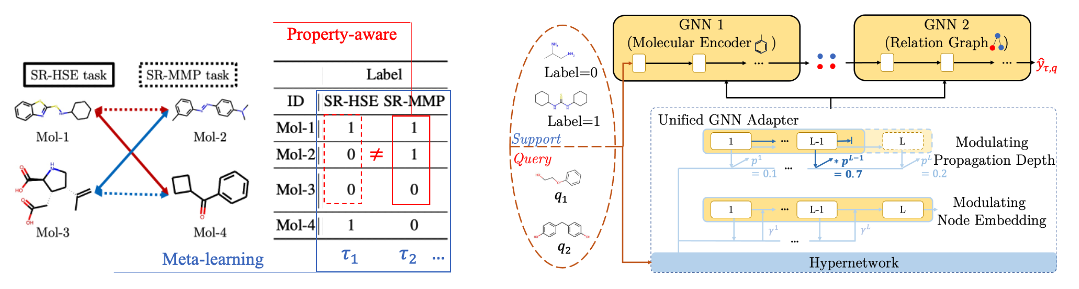}
	
	\vspace{-10pt}
	\caption{PAR meta-learns a property-aware network to improve sample efficiency (left). 
		PACIA introduces GNN adapter to achieve efficient hierarchical adaptation (right).}
	\vspace{-20pt}
	\label{fig:par}
\end{figure*}

\begin{table}[t]
	\renewcommand{\arraystretch}{1.50}
	\centering
	\small
	\begin{tabular}{c|c|c|c}
		\hline
		Model & WN18RR & FB15k237 & YAGO3-10 \\
		\hline
		Interstellar & 0.48 & 0.32 & 0.51 \\
		CompGCN & 0.479 & 0.355 & 0.421 \\
		AutoBLM & 0.490  &  0.360  & 0.571  \\
		\hline
	\end{tabular}
	\captionof{table}{MRR performance comparison of AutoBLM and other neural models.}
	\label{tab:autoblm}
	\vspace{-20pt}
\end{table}

\subsection{Symbolized Architecture Search for Recommendation}

User cold-start recommendation problem targets at quickly generalize to new tasks (i.e. personalized recommendation for cold-start users) with a few training samples (i.e. a few interaction histories).  
A number of works~\citep{dong2020mamo,lee2019melu,lu2020meta} adopt the classic gradient-based meta-learning strategy called model-agnostic meta learning (MAML)~\citep{finn2017model}, 
which learns a good initialized parameter from a set of tasks and adapts it to a new task by taking a few steps of gradient descent updates on a limited number of labeled samples. 
This line of models has demonstrated high potential of alleviating user cold-start problem. 
However, gradient-based meta-learning strategy require expertise to tune the optimization procedure to avoid over-fitting. 
Besides, the inference time can be long.
To address these challenges, we propose ColdNAS~\citep{Wu2023}, 
which searches for proper modulation structures to adapt user-specific recommendations as shown in Figure~\ref{fig:coldnas}. 
The core of ColdNAS is to utilize a hypernetwork that maps user interaction history to personalized parameters, which are then used to modulate the predictor model. Specifically, ColdNAS formulates a unified search space and optimizes the modulation functions as part of a differentiable neural architecture search (NAS). 
By adopting 
a symbolic search approach instead of deep learning-based fitting methods, 
ColdNAS significantly reduces the search space and improves efficiency. 
Furthermore, models that are better suited for specific datasets can be learned through an efficient and robust search algorithm, 
achieving adaptability across datasets.

The performance of ColdNAS is evaluated on multiple benchmark datasets 
(i.e., MovieLens~\citep{harper2015movielens}, Book-Crossing~\citep{ziegler2005improving} and Last.fm~\citep{lin2021task}), 
and measured with metrics such as MSE and MAE, 
are shown in the 
Table~\ref{tab:coldnas}. Compared with existing state-of-the-art cold-start models 
(i.e., MAML and TaNP~\citep{lin2021task}) demonstrates significant improvements on all datasets. 
In particular, it automatically identifies the optimal modulation structures that outperform fixed modulation strategies, 
yielding superior performance while maintaining computational efficiency.

\begin{table}[t]
	\renewcommand{\arraystretch}{1.50}
	\centering
	\small
	\begin{tabular}{p{1.0cm}|p{0.5cm}|p{1.1cm}|p{1.1cm}|p{1cm}}
		\hline
		Dataset                    & Metric & MAMO             & TaNP             & ColdNAS          \\ \hline
		\multirow{2}{*}{MovieLens} & MSE    & $90.20_{(0.22)}$ & $89.11_{(0.18)}$ & \textbf{87.96$_{(0.12)}$}\\
		& MAE    & $75.34_{(0.26)}$ & $74.78_{(0.14)}$ & \textbf{74.29$_{(0.20)}$} \\ \hline
		Book                       & MSE    & $14.82_{(0.05)}$ & $14.75_{(0.05)}$ & \textbf{14.15$_{(0.08)}$} \\
		Crossing                   & MAE    & $3.51_{(0.02)}$  & $3.48_{(0.01)}$  & \textbf{3.40$_{(0.01)}$}  \\ \hline
		\multirow{2}{*}{Last.fm}   & MSE    & $21.64_{(0.10)}$ & $21.58_{(0.20)}$ & \textbf{20.91$_{(0.05)}$} \\
		& MAE    & $42.30_{(0.28)}$ & $42.15_{(0.56)}$ & \textbf{41.78$_{(0.24)}$} \\ \hline
	\end{tabular}
	\captionof{table}{Test performance (\%) obtained on benchmark datasets. The best results are highlighted in bold and the second-best in italic. A smaller value is better.}
	\label{tab:coldnas}
	\vspace{-10pt}
\end{table}

\section{Parsimony on Training}
\label{efficiency}

In AI-assisted scientific research, particularly in molecules and biomedicine, 
the scarcity of labeled data presents significant challenges. 
Motivated by~\citep{palatucci2009zero,rastogi2022learning,ilharco2022editing},
which utilizes the symbolic structure between parameters obtained from different tasks for downstream tasks prediction,
we propose approaches to learning parsimony by leveraging task-related knowledge to guide the fine-tuning on downstream tasks. 
In 
molecules properties prediction,
we have developed meta-learning learning techniques
that enforce parsimony on learning. 
These approaches ensure that parameters can be efficiently adapted in relation to functional groups, 
optimizing their use and enhancing model adaptability. 

\subsection{Property-Aware Relation Networks}

Our first work towards this problem is the property-aware relation networks (PAR) \citep{wang2021property,yao2024property}. 
PAR uses a property-aware molecular encoder to transform the generic molecular embeddings to property-aware ones.
To fully leverage the supervised learning signal, PAR learns  to estimate the
molecular relation graph by a query-dependent relation graph learning module, in
which
molecular embeddings are refined w.r.t. the target property. 
Thus, 
the facts that both property-related information
and relationships among molecules change across different properties
are utilized to better learn and propagate molecular embeddings. 
Besides, 
we propose a selective update strategy for handling generic and property-aware information. In the inner-loop update, only the property-aware information is updated, while both generic and property-aware information are updated simultaneously in the outer-loop.
We use gradient descents to update parameters in both loops. 
Through the selective update strategy, 
the model can capture generic and property-aware information separately in the training procedure.

\begin{table}[t]
\center
\small
\setlength\tabcolsep{10pt}
\begin{tabular}{c|c|c|c|c|c|c}
	\hline
	& PACIA & \multicolumn{4}{c}{PAR} \\\hline
	\# Total para.&3.28M& \multicolumn{4}{c}{2.31M}\\
	\# Adaptive para.&3.00K& \multicolumn{4}{c}{0.38M}\
 \\\hline
\end{tabular}
\caption{Total and adaptive parameters of PACIA and PAR in few-shot learning.}
\label{tab:adapt-new}
\end{table}

The results are reported in Table~\ref{tab:fsl-results}. From the results, we observe that PAR obtains the best performance among methods using graph-based molecular encoders learned from scratch.
The outperforming results can be attributed to the combination of metric-based and optimization-based method in the design of
PAR method. In terms of average improvement, PAR obtains
significantly better performance than the best baseline learned
from scratch (e.g. EGNN) by 1.59\%, showing enhanced performance.

\subsection{Parameter-Efficient GNN Adapter}
We further introduce parameter-efficient graph neural network (GNN) adapter (PACIA) \citep{wu2024pacia}. 
By adopting this approach, PACIA significantly reduces the risk of overfitting. Moreover, it offers the advantage of faster inference speeds, as the adapted parameters are generated through a single forward pass rather than through iterative optimization steps. Additionally, initializing the function $g(\cdot)$ with neural networks allows for more flexible forms of updates compared to traditional gradient descent methods.  
Further,  
we 
design a hierarchical 
adaptation mechanism in the framework: 
Task-level adaptation is achieved in the encoder since the structural features in
molecular graphs needs to be captured  in a property-adaptive manner, while
query-level adaptation is achieved in the predictor based on the property-adaptive representations.  
To adapt GNN's parameter-efficiently, 
we design a hypernetwork-based GNN adapter to 
generate a few adaptive parameters
to modulate the node embedding and propagation depth, 
which are essential in message passing process.  
No further fine-tuning is required.

\begin{table*}[t]
	\centering
	\small
	\renewcommand{\arraystretch}{1.50}
	\setlength\tabcolsep{2pt}
	\begin{tabular}{c|cc|cc|cc|cc}
		\hline
		\multirow{2}{*}{Method}  &\multicolumn{2}{c|}{Tox21} &\multicolumn{2}{c|}{SIDER}
		& \multicolumn{2}{c|}{MUV}&\multicolumn{2}{c}{ToxCast} \\
		&10-shot&1-shot&10-shot&1-shot&10-shot&1-shot&10-shot&1-shot\\\hline
		MAML&$79.59_{(0.33)}$&$75.63_{(0.18)}$&$70.49_{(0.54)}$&$68.63_{(1.51)}$&$68.38_{(1.27)}$&$65.82_{(2.49)}$&$68.43_{(1.85)}$&$66.75_{(1.62)}$\\
		IterRefLSTM&$81.10_{(0.10)}$&$80.97_{(0.06)}$&$69.63_{(0.16)}$&$71.73_{(0.06)}$&$49.56_{(2.32)}$&$48.54_{(1.48)}$&-&-\\
		\textbf{PAR}&$82.13_{(0.26)}$&$\underline{80.02}_{(0.30)}$&$\underline{75.15}_{(0.35)}$&$\underline{72.33}_{(0.47)}$&$68.08_{(2.23)}$&$65.62_{(3.49)}$&$70.01_{(0.85)}$&$\underline{68.22}_{(1.34)}$\\
		ADKF-IFT&$\underline{82.43}_{(0.60)}$&$77.94_{(0.91)}$&$67.72_{(1.21)}$&$58.69_{(1.44)}$&$\bm{98.18}_{(3.05)}$&$\underline{67.04}_{(4.86)}$&$\underline{72.07}_{(0.81)}$&$67.50_{(1.23)}$\\
		\textbf{PACIA}&$\bm{84.25_{(0.31)}}$&$\bm{82.77_{(0.15)}}$&$\bm{82.40_{(0.26)}}$&$\bm{77.72_{(0.34)}}$&$\underline{72.58}_{(2.23)}$&$\bm{68.80}_{(4.01)}$&$\bm{72.38_{(0.96)}}$&$\bm{69.89_{(1.17)}}$\\
		\hline
	\end{tabular}
	
	\caption{Test ROC-AUC (\%) obtained on MoleculeNet. 
		The best results 
		are bolded, second-best results are underlined. }
	\label{tab:fsl-results}
	\vspace{-20pt}
\end{table*}

Table~\ref{tab:fsl-results} shows the prediction performance.  
Our analysis reveals that methods utilizing pretrained graph-based molecular encoders generally outperform those with encoders learned from scratch. 
This underscores the effectiveness of pretrained encoders in capturing rich, 
generic molecular information, subsequently providing superior molecular embeddings. 
In further evaluations, meta-learning methods that learn relational graphs—specifically EGNN, PAR, and PACIA—demonstrate enhanced performance. 
Notably, PACIA consistently achieves the highest ROC-AUC scores, followed closely by PAR. 
In summary, PACIA establishes itself as the new state-of-the-art for predicting molecular properties. 
Additionally, as shown in Table~\ref{tab:adapt-new}, PACIA can obtain better performance due to the reduction of adaptive parameters, which also leads to better generalization and alleviates the risk of overfitting to a few shots. 

\begin{figure}[t]
	\centering
	\includegraphics[width=1\linewidth]{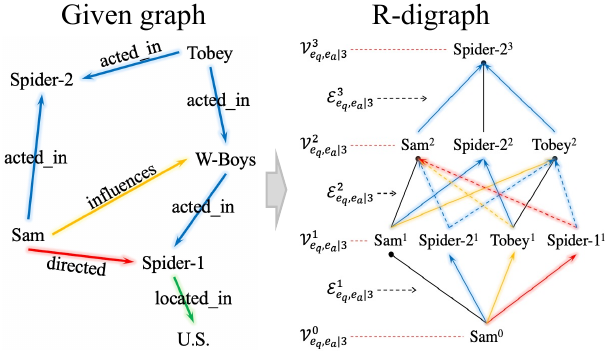}
	\caption{RED-GNN makes use of dynamic programming to
		recursively encodes multiple r-digraphs with shared edges, and
		utilizes query-dependent attention mechanism to select the strongly
		correlated edges.}
	\label{fig:redgnn}
	\vspace{-10pt}
\end{figure}

\section{Parsimony on Interpretation}
\label{interpretability}

Interpretability is important for understanding the result. 
In many fields, experts need to interpret and understand the results of models in order to make informed decisions. 
GNNs, though effective in learning from relational graphs, is very challenging to provide interpretable inference when facing a massive amount of associated information on graphs.
For graph learning,
the core problem in achieving model interpretability lies in accurately capturing strong logical relationships and 
exhibiting inference process with evidence. 
To achieve interpretability on graphs,
our key idea is to 
capture the logical rules
inside graphs as supporting evidence
and 
use the logical interpretation of knowledge to clarify models' reasoning process.
In the following,
we introduce the idea of subgraph learning 
for interpreting
and show the interpretable inference process with the learned subgraphs.

\subsection{Interpreting with Subgraph Learning}
The relational graph structures are complex and hard to understand directly.
In comparison,
relational paths 
with multiple connected edges are more interpretable
\citep{xiong2017deeppath,das2018go}.
Based on this observation,
RED-GNN \citep{zhang2022knowledge}
introduces a new relational structure, called relational di-graph (r-digraph) as illustrated in Figure~\ref{fig:redgnn}.
The r-digraphs generalize relational paths to subgraphs by preserving the overlapped relational paths
and structures of relations for reasoning.
By leveraging the GNN model with attention mechanism
to propagate information over the subgraph,
significant performance improvement has been achieved 
and logical paths in r-digraphs can be captured.

\begin{table}[t]
	\small
	\renewcommand{\arraystretch}{1.50}
	\setlength\tabcolsep{1pt}
	\begin{tabular}{c|ccc|ccc|ccc}
		\hline
		\multirow{2}{*}{Model} & \multicolumn{3}{c|}{WN18RR} & \multicolumn{3}{c|}{FB15k-237} & \multicolumn{3}{c}{YAGO3-10}  \\
		& MRR & H@1 & H@10 & MRR & H@1 & H@10  & MRR & H@1 & H@10 \\
		\hline
		DRUM & 0.486 & 42.5 & 58.6 &  0.343 & 25.5 & 51.6		& 0.531  & 45.3 & {67.6} \\
		RNNLogic & 0.483 & 44.6  & 55.8  & 0.344 & 25.2 & 53.0		& \underline{0.554} & \textbf{50.9}  &  67.3 \\
		\hline
		CompGCN & 0.487 & 44.3 & 54.6 & 0.355 & 26.4 & 53.5 		 & 0.421 & 39.2  & 57.7 \\
		\hline
		NBFNet & \underline{0.551}  & \underline{49.7}  & \underline{66.6} 	& \underline{0.415} & \underline{32.1} & \textbf{59.9}	&  0.550 & 47.9 & 68.6 \\
		RED-GNN & \textbf{0.564} & \textbf{50.2} & \textbf{67.8} & \textbf{0.418} & \textbf{32.9} & \underline{59.0}	  & \textbf{0.584} & \textbf{50.9} & \underline{71.3} \\
		\hline
	\end{tabular}
	\caption{Performance comparison on knowledge graph reasoning tasks. The best results are bolded and the second-best underlined.}
	\label{tab:redgnn}
	\vspace{-10pt}
\end{table}

\begin{figure}[t]
	\centering
	\includegraphics[width=0.90\linewidth]{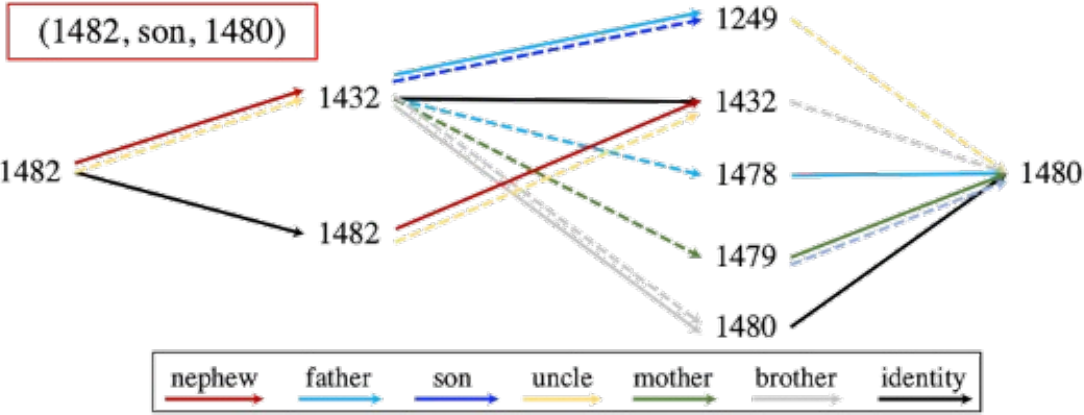}
	\caption{Visualization of the learned structures. Dashed lines
		mean inverse relations. The query triples are indicated by
		the red rectangles.}
	\label{fig:case-redgnn}
	\vspace{-10pt}
\end{figure}

The effectiveness of subgraph learning methods are evaluated on 
general knowledge graph reasoning benchmarks.
As shown in Table~\ref{tab:redgnn},
the subgraph learning methods NBFNet \citep{zhu2021neural}
and RED-GNN \citep{zhang2022knowledge} show significant advantage
over the embedding-based methods. 
We visualize an exemplar learned r-digraphs by RED-GNN on the Family dataset.
Figure~\ref{fig:case-redgnn} shows one
triple that DRUM fails.
Chain-like structures alone, such as 
\texttt{id-1482}
$\xrightarrow{\text{\tiny nephew}}$ 
\texttt{id-1432} 
$\xrightarrow{\text{\tiny brother}}$ 
\texttt{id-1480} 
$\xrightarrow{\text{\tiny identify}}$ 
\texttt{id-1480} 
or 
\texttt{id-1482 }
$\xrightarrow{\text{\tiny nephew}}$ 
\texttt{id-1432} 
$\xrightarrow{\text{\tiny father}}$ 
\texttt{id-1249} 
$\xrightarrow{\text{\tiny uncle}}$ 
\texttt{id-1480}, 
can only imply that 
\texttt{id-1482} 
is the son or nephew of 
\texttt{id-1480}.
These two chain-like structures together provide the evidence that 
\texttt{id-1432} 
and 
\texttt{id-1480} are the only brother of each other, 
which is crucial in inferring that 
\texttt{id-1482} 
is the son of 
\texttt{id-1480}. 

\begin{figure}[t]
	\centering
	\includegraphics[width=1\linewidth]{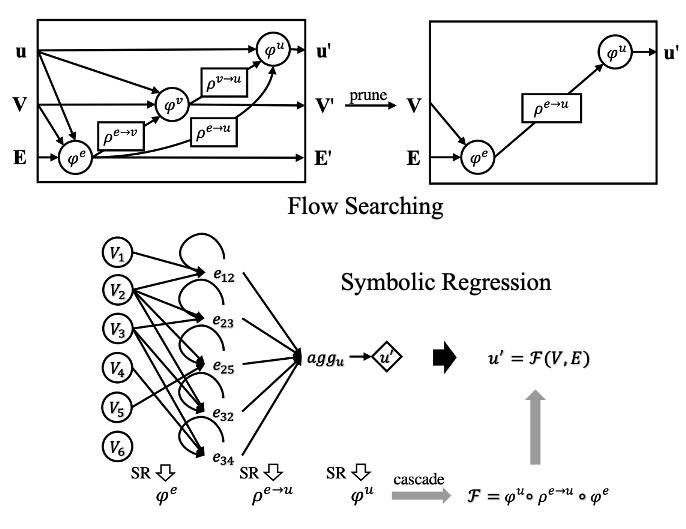}
	\caption{GSR models the formula skeleton
		with a message-passing flow, which helps transform the discovery of the skeleton
		into the search for the message-passing flow. Then, the 
		formulas can be identified by interpreting component functions of the searched
		message-passing flow, reusing classical symbolic regression methods.}
	\label{fig:gsr}
	\vspace{-10pt}
\end{figure}

\begin{figure}[t]
	\centering
	\vspace{-10pt}
	\includegraphics[width=1\linewidth]{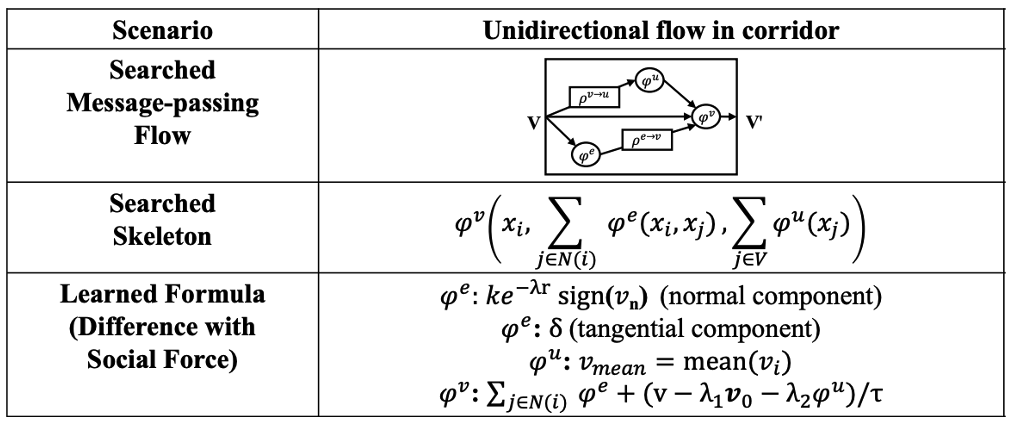}
	\caption{Learned formulas for pedestrian dynamics.}
	\label{fig:case-gsr}
	\vspace{-10pt}
\end{figure}

\begin{figure*}[t]
\centering
\includegraphics[width=1\linewidth]{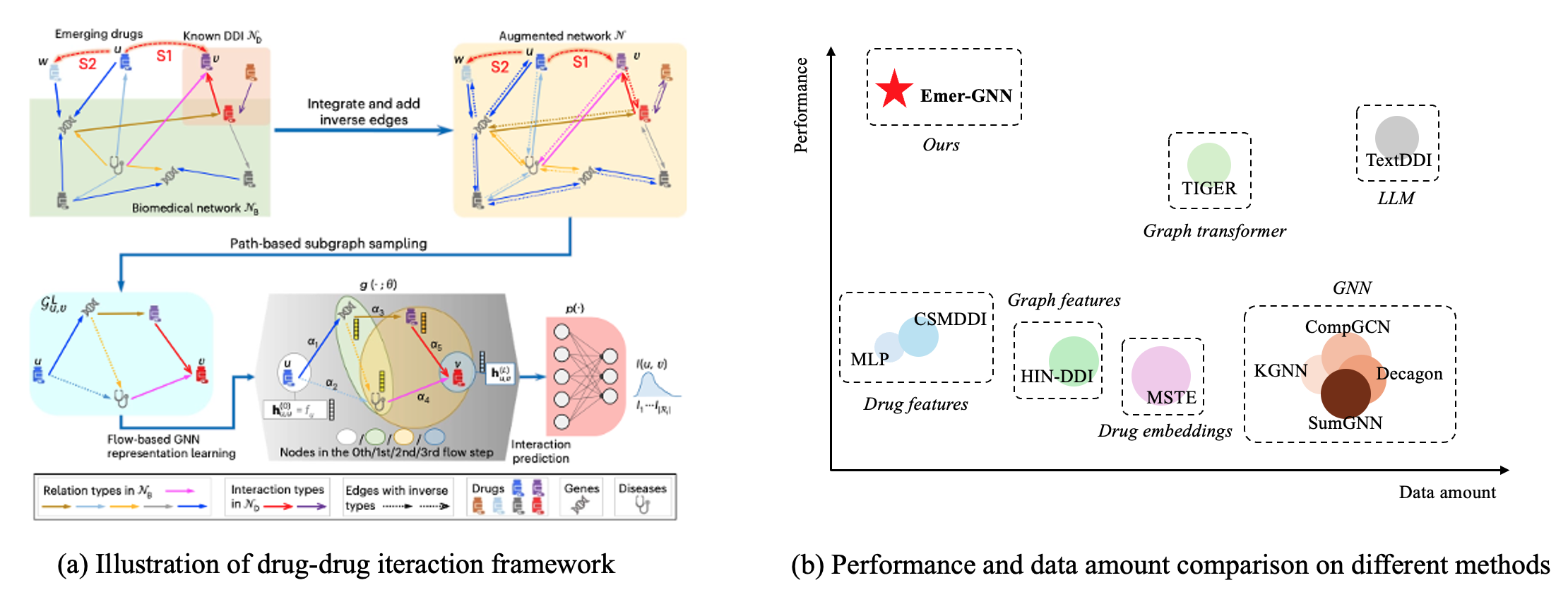}

\vspace{-5pt}
\caption{(a) EmerGNN learns pairwise representations of drugs by extracting the paths between drug pairs, 
	propagating information from one drug to the other,
	and incorporating the relevant biomedical concepts on the paths. (b) Performance and data amount comparison between traditional, transformer, and LLM-based methods.}
\label{fig:emergnn}
\vspace{-20pt}
\end{figure*}

\begin{figure*}[t]
\centering
\includegraphics[width=0.8\linewidth]{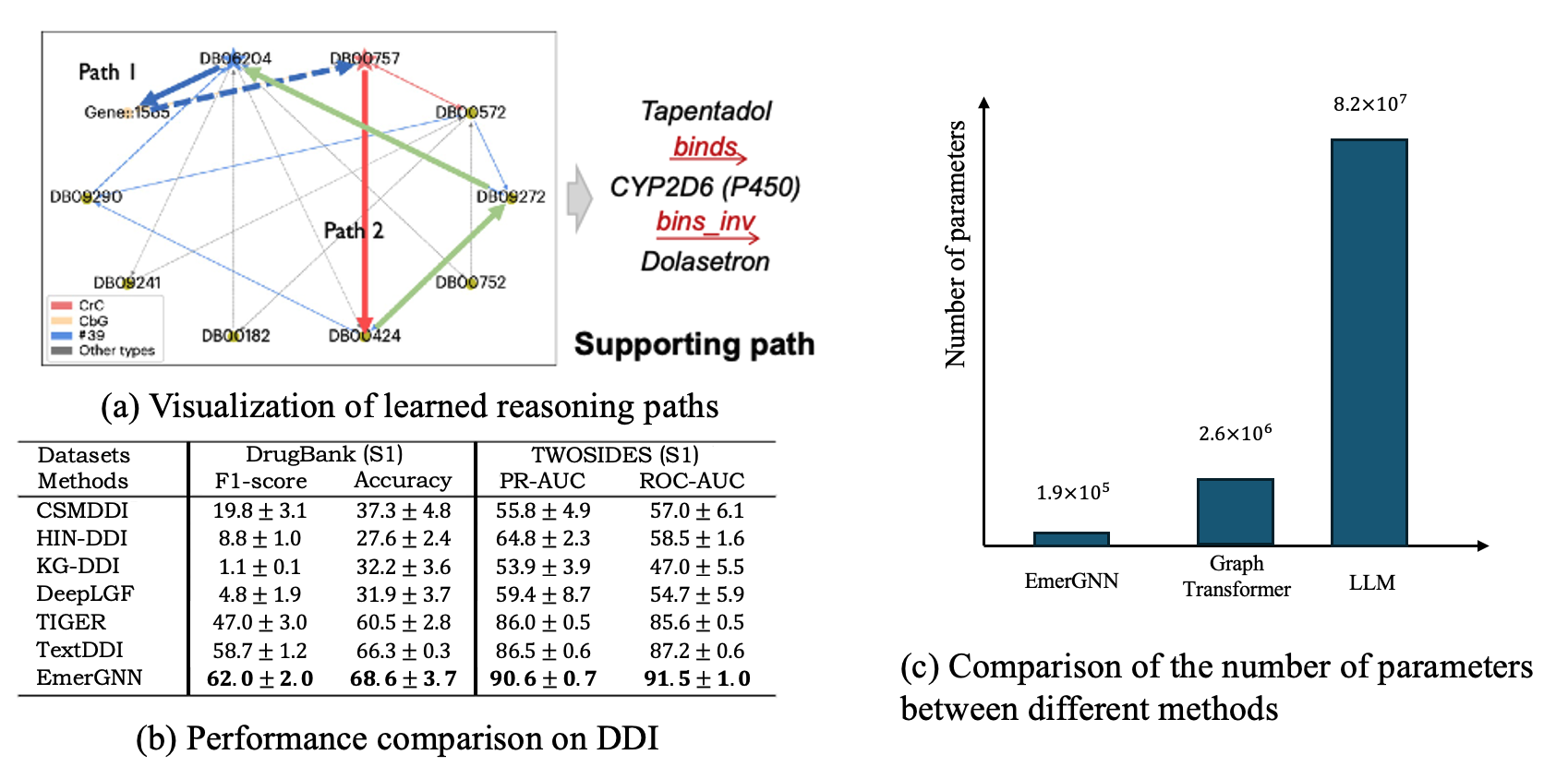}
\vspace{-5pt}
\caption{(a) Visualization of the structure learned by EmerGNN.
	DB006204 (Tapentadol) is an existing drug,
	and DB00757 (Dolasetron) is an emerging drug. (b) Performance comparison on DDI prediction tasks.
	The best results are bolded and the second-best underlined. (c) The number of parameters comparison between EmerGNN, Graph Transformer, and LLM-based methods.}
\label{fig:parameter}
\vspace{-20pt}
\end{figure*}

\subsection{Symbolic Regression on Graphs}

Graph-structured physical mechanisms are commonly found in various scientific domains, 
where variables such as mass, force, 
and energy interact through relationships on a graph. 
Traditional symbolic regression (SR) methods have been used to discover formulas from input-output data pairs but struggle to handle graph-structured inputs. 
We present the GSR~\citep{shi2023learning} as shown in Figure~\ref{fig:gsr}, which is designed to extend symbolic regression to graph-structured physical mechanisms. By integrating prior knowledge in the form of symbolic logic and physical laws, the proposed approach transforms the discovery of formula skeletons into a search for efficient message-passing flows in GNNs. The framework enables the model to balance accuracy and simplicity by identifying Pareto-optimal flows that generalize well across different physical domains. 
Through knowledge-guided symbolic learning, 
the model captures essential relationships within graph-structured data, achieving parsimony on interpretation.

The learned formulas and the corresponding physical meanings are reported in
Figure~\ref{fig:case-gsr}, which demonstrates that our model can learn different skeletons and formulas that are more
precise than the social force model with explicit physical meanings.

\section{Potential in Drug Development}

In the ever-changing world of pharmaceuticals, 
the intersection of scientific advancements and regulatory changes has led to a significant breakthrough, 
particularly in the rapid development of new drugs aimed at treating rare, severe, 
or life-threatening diseases~\citep{su2022trends,ledford2022hundreds}. 
Machine learning models
have become powerful tools for predicting drug interactions, 
capable of comprehensively capturing intricate relationships on the interaction graphs. 
Our work EmerGNN \citep{zhang2023emerging} takes this a step further by introducing a graph learning framework that predicts interactions between emerging and existing drugs by integrating drug-drug interaction (DDI) and biomedical networks, extracting path-based subgraphs,
and learning subgraph representations.

The general idea of EmerGNN is illustrated in 
Figure~\ref{fig:emergnn}(a).
Emerging drugs typically have limited interactions with existing drugs. 
To address this issue, 
EmerGNN utilizes biomedical network and extracts subgraph 
$\mathcal G_{u,v}^L$
from it 
to connect emerging drug $u$ with existing drug $v$. 
Drawing inspiration from the subgraph learning methods \citep{zhang2022knowledge,zhu2021neural}, 
EmerGNN proposes a flow-based GNN 
$g(\mathcal G_{u,v}^L;\bm \theta)$
with an attention mechanism to encode pair-wise subgraph representations for drug pairs. 
These subgraph representations are then used to directly predict the interaction of the drug pair. 
This approach enables a more nuanced understanding of interactions involving emerging drugs, 
incorporating knowledge from the interconnected network of biomedical entities and their relationships, 
and providing interpretable insights.
We also compare the performance and data requirements of traditional, transformer-based, and LLM-based methods, with the results shown in Figure~\ref{fig:emergnn}(b). By learning pairwise representations of drugs, EmerGNN achieves superior performance while using less data compared to other methods.

The performance comparison on DDI prediction tasks is shown in 
Figure~\ref{fig:parameter}(b). From the results, we find that the proposed
EmerGNN outperforms 
CSMDDI \citep{liu2022predict} that directly uses drug features to predict,
HINDDI \citep{tanvir2021predicting} that counts meta-paths from biomedical network for prediction,
KG-DDI \citep{karim2019drug} that learns drug embeddings,
and DeepLGF \citep{ren2022biomedical} that models with GNN
in both benchmarks and both settings,
TIGER~\citep{su2024dual} that leverages the
Transformer architecture to effectively exploit the structure of heterogeneous graph.
TextDDI~\citep{DBLP:conf/emnlp/ZhuZC0X23} uses a language model-based DDI predictor and a reinforcement learning (RL)-based information selector.
Furthermore,
we show a case study with the visualized subgraph in Figure~\ref{fig:parameter}(a)
by selecting important paths according to the attention values.
The path connecting two drugs through the binding protein \texttt{Gene::1565 (CYP2D6)}, 
which is a \texttt{P450} enzyme that plays a key role in drug metabolism,
for example,
shows a way to interpret the inference process with supporting evidence.
From the empirical comparison and case studies,
we conclude that subgraph learning can not only achieve improvements in representation learning,
but also interpret the inference process with supporting evidence.
Additionally, we compare the number of parameters across different methods, with the results shown in Figure~\ref{fig:parameter}(c). By incorporating biomedical relationships between subgraphs and drugs within the knowledge space, we effectively reduce the number of training parameters while maintaining superior performance.

\section{Future Works}

Finally,
we talk about our ongoing future works 
from theory,
method and application.
\begin{itemize}
\item 
\textit{Theory.}
Current machine learning relies on large-scale data and complex models, but distilling knowledge into compact forms can make models more efficient. Future work will quantify the necessary domain-specific knowledge to enhance model performance while reducing large datasets. This involves formalizing ``knowledge as a resource'' and creating algorithms to optimize knowledge use while maintaining performance across tasks. A theoretical framework for knowledge quantization will enable better integration of symbolic logic, physical laws, and structured knowledge into learning models.

\vspace{5px}

\item 
\textit{Methods.}
As LLMs dominate AI research, integrating them with specialized knowledge and smaller models is crucial. Future work will focus on methods for seamless integration of LLMs with domain-specific frameworks. Instead of indefinitely scaleup models, the focus will be on using LLMs as modular components that can be fine-tuned based on task-specific knowledge. This approach involves developing techniques to align LLM outputs with symbolic and logical reasoning systems, enhancing performance in specialized fields like bioinformatics or physics. Additionally, reducing the computational footprint of LLMs while maintaining their versatility will be a key focus.

\vspace{5px}

\item 
\textit{Applications.}
AI's application in scientific discovery is highly promising. Future research will use parsimonious learning to address complex challenges in drug discovery, molecular property prediction, and other scientific areas. In drug-drug interaction prediction, simplifying neural networks can reduce computational demands. For protein structure prediction, leveraging domain knowledge can improve accuracy despite limited data. In molecular property prediction, parsimony enhances model transparency and trustworthiness by extracting key insights from complex data.
\end{itemize}

Through endeavor in above directions,
we aim to develop a unified and general methodology where knowledge representations and functional spaces co-evolve synergistically. This approach involves designing frameworks that leverage task-specific domain knowledge to 
\textcolor{black}{simultaneously reduce model complexity, minimize training parameters, and enhance interpretability. }
The goal is for knowledge to refine the functional space — such as model architectures and representations — while the learned functions provide feedback to enrich and expand the knowledge space. 
This unified methodology promises to create more adaptable, interpretable, 
and efficient learning systems.

\section{Conclusion}

This paper introduces 
an alternative way to develop next-generation learning techniques instead of brute-force scaleup.
By leveraging the duality between data and knowledge, 
our method extracts symbolic logic and physical laws during the learning process and applies combinatorial generalization to various tasks. This approach effectively overcomes the limitations of traditional scaling methods. 
Experimental results demonstrate that our framework significantly improves model performance, 
showcasing its ability to achieve parsimony on model, training and interpretation. 
These findings underscore the potential of integrating knowledge into machine learning models, 
offering a promising direction for future research and applications.

\section*{Acknowledgment}

Q. Yao work is supported by National Key Research and Development
Program of China (under Grant No.2023YFB2903904), National
Natural Science Foundation of China (under Grant No.92270106)
and Beijing Natural Science Foundation (under Grant No.4242039).
J. Kwok is supported in part by the Research Grants Council of 
the Hong Kong Special Administrative Region (Grants 16202523 and C7004-22G-1).
Finally,
we thanks suggestive comments 
from reviewers, Wenguang Chen and Xuefei Ning
for helping us improve the quality of this perspective.

\section*{Conflict of Interest}
The authors have no conflicts of interest to report.


\bibliographystyle{plainnat}
\bibliography{AAAI}

\section{Author Biographies}

Quanming Yao is a tenure-track assistant professor at the Department of Electronic Engineering, Tsinghua University. He obtained his Ph.D. degree at the Department of Computer Science and Engineering of Hong Kong University of Science and Technology (HKUST). He regularly serves as area chairs for ICML, NeurlPS and ICLR. He is also a receipt of National Youth Talent Plan (China), Forbes 30 Under 30 (China) Young Scientist Awards (Hong Kong Institution of Science), and Google Fellowship (in machine learning).

Yongqi Zhang is currently a tenure-track assistant professor in HKUST(GZ). Prior to this, he worked as a research scientist at the AI listed company 4Paradigm Inc from 2020 to 2023. He completed his doctoral degree in Computer Science and Engineering from HKUST. 
His research interest lies in large language models and knowledge graphs. He has published more than 20 papers on top tier venues, including Nature Computational Science, IEEE TPAMI, KDD, NeurIPS, etc.

Yaqing Wang is currently an Associate Professor at the Beijing Institute of Mathematical Sciences and Applications. She received her Ph.D. in Computer Science and Engineering from the Hong Kong University of Science and Technology. 
Dr. Wang has published over 20 papers in top-tier international conferences and journals, including NeurIPS, ICML, KDD, SIGIR, TheWebConf, IJCAI, TPAMI, JMLR, and TIP, with more than 4000 citations. Dr. Wang has been selected for the World’s Top 2\% Scientists List (single year) by Stanford University and Elsevier company on 2024.

Nan Yin is currently a research associate at the Department of Computer Science and Engineering of Hong Kong of Computer Science and Technology. He received the B.S. degree from National University of Defense Technology, Changsha,~China, in 2016 and the Ph.D. degree in the School of Computer Science and Technology, National University of Defense Technology, Changsha, China. His current research interests include transfer learning and graphs.

James Kwok 
is currently a Professor with the Department of Computer Science and Engineering,
Hong Kong University of Science and Technology.
He is serving / served as an Associate Editor for the IEEE Transactions on Neural Networks and Learning Systems, Neural Networks, Neurocomputing, Artificial Intelligence Journal, and on the Editorial Board of Machine Learning. 
He is also serving as Senior Area Chairs of major machine learning / AI conferences including NeurIPS, ICML, ICLR, IJCAI, and as Area Chairs of conferences including AAAI and ECML. He is on the IJCAI Board of Trustees. 
He is recognized as the Most Influential Scholar Award Honorable Mention for "outstanding and vibrant contributions to the field of AAAI/IJCAI between 2009 and 2019". Prof Kwok is the IJCAI-2025 Program Chair.

Qiang Yang is a fellow of Canadian Academy of Engineering (CAE) and Royal Society of Canada (RSC), Chief Artificial Intelligence Officer of WeBank, a chair professor of Computer Science and Engineering Department at Hong Kong University of Science and Technology (HKUST). 
He is also a fellow of AAAI, ACM, CAAI, IAPR, and AAAS. 
He was the Founding Editor in Chief of the ACM Transactions on Intelligent Systems and Technology (ACM TIST) and the Founding Editor in Chief of IEEE Transactions on Big Data (IEEE TBD). He received the ACM SIGKDD Distinguished Service Award, in 2017 and the Wu Wenjun outstanding contribution award of artificial intelligence, in 2019. His research interests are artificial intelligence, machine learning, data mining, and planning. He had been the Founding Director of the Huawei’s Noah’s Ark Research Lab between 2012 and 2015, the Founding Director of HKUST’s Big Data Institute, the Founder of 4Paradigm and the President of IJCAI (2017-2019). His latest books are Transfer Learning, Federated Learning, Privacy-preserving Computing, and Practicing Federated Learning.

\end{document}